# Logical Evaluation of Consciousness: For Incorporating Consciousness into Machine Architecture

Mr C.N.Padhy, Ms. R.R.Panda
**Institute of Knowledge and Information Technology(IKIT)**

**Abstract**
Machine Consciousness is the study of consciousness in a biological, philosophical, mathematical and physical perspective and designing a model that can fit into a programmable system architecture. Prime objective of the study is to make the system architecture behave consciously like a biological model does. Present work has developed a feasible definition of consciousness, that characterizes consciousness with four parameters i.e., parasitic, symbiotic, self referral and reproduction. Present work has also developed a biologically inspired consciousness architecture that has following layers: quantum layer, cellular layer, organ layer and behavioral layer and traced the characteristics of consciousness at each layer. Finally, the work has estimated physical and algorithmic architecture to devise a system that can behave consciously.

## 1. Origin of Consciousness

How biological being are conscious? How this consciousness is organized? are few of the questions that are attempted in this section, by taking into consideration of various views from philosophy, physical science, biological science and mathematics. It clearly depict that, consciousness is the agent that is responsible for intelligent behavior or phenomenon. Now we will discuss about the various conceptualizations about consciousness.

Many phenomenon in the universe are not sufficiently explained by the scientific studies, but aptly explained by philosophy. Hence study of philosophy many a times gives much impetus to probe into the causality of the phenomenon and finally develop the model so as to apply for solving real life problems.

Here our objective is to induct consciousness into the machine architecture, so that intelligence aspect of machine can be autonomous as in case of any living being. Even an ant is autonomously intelligent and do not require any external control for it's survival or normal functioning.

It is essential to probe into the causality of consciousness and logical schemata of consciousness for building a model inspired by this phenomenon. Once, the model is developed, it can be suitably fit into programmable machine architecture and can behave consciously. Here, the author have attempted to probe into the philosophy of consciousness, but found that frame of creation is highly required for understanding the concept of consciousness. Hence, I am depicting here the framework of creation for better appreciation of phenomenon of consciousness.

### 1.2 Framework of Creation
The whole creation is manifested through a well organized layers, that starts with the formless characteristics universe and ends at intelligent actions generated in a matter.
All the layers have shown in the Fig 2. and narrated in the followed texts:

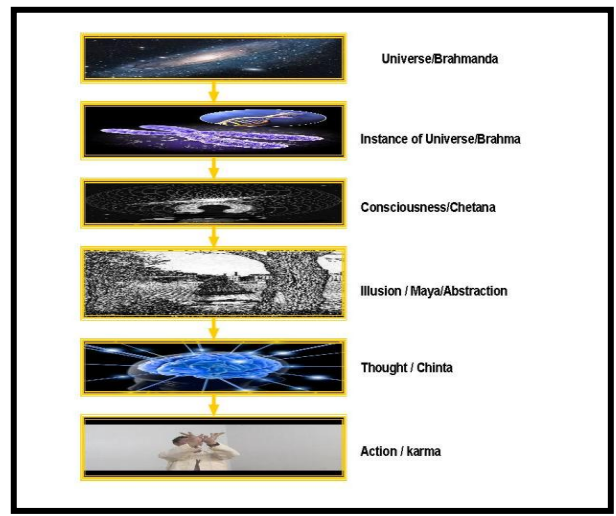

### 1.2.1 Universe / Brahmanda
The whole creation is manifested and characterized by Universe(Brahmanda). As per the Oxford Dictionary "all existing matter and space considered as a whole; the cosmos" and as per Webster Dictionary "the whole body of things and phenomena observed or postulated". Universe or Brahmanda is the *prakriti* or set of characteristics of whole matter and phenomenon, that exists in this universe. In brief, Universe is composed of

- Matter,
- Space, and
- Knowledge Base.

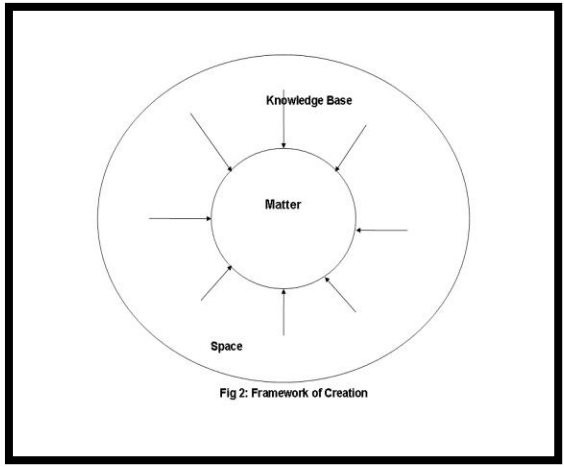

**Knowledge Base**

Knowledge base is the well defined, logically arranged set of instructions for execution of matters and various phenomenon..

**Space**

Space is the universe of discourse or domain within which knowledge base is incident on matter.

**Matter**

Any thing that governs with the universal knowledge base and within a space of enactment is a matter.

**Matter = Space + Knowledge Base**

**1.2.2 Brahman :**
- Brahman is the instance of Brahmanda,
- Brahman has no such difference as it is one without a second,
  - Brahman is indivisible as it is devoid of all differences, whether within itself, or from things of same clsss or from things of other classes(Veda)
  - It is formless in essence(Veda)
  - Before creation there was only Brahman which is existence itself on which whole creation is superimposed and percepted as existent(Chandogya Upanishad)

**Interpretation of Brahman**
- Brahman is wide spread and exists in all matter
- Brahman is formless in essence and when Brahman is incident on matter it gets the form,
- Brahman is responsible for consciousness in the matter
- Characteristics of Universe is also exists in each matter in the form of Brahman, hence they are coherence with the Brahmanda or Universe.

**1.2.3 Consciousness(Chetana)**
- Consciousness is the is the incidence of Brahman in the matter, that creates the *Prakriti* or Characteristics of a matter.
- Consciousness is the layer in the framework of creation that is responsible for executing the universal knowledge base in a matter.

**Levels of Consciousness**

Consciousness can be classified into following levels:
- Quantum Layer
- Compound Layer
- Cellular Layer
- Organ Layer
- Behavioral Layer

At each lavel consciousness can execute the matter autonomously and can also gives rise to next layer of consciousness to create a higher level of matter. In fact consciousness is responsible for generating the feel of universe at matter, as well as feel of matter in the universe, hence creating the coherency between universe and matter. Following figure depicts the various levels of consciousness.

In this work, detail narration Cell level consciousness, Organ level and Behavioral Level is given. But, as currently work is in progress for quantum, compound level consciousness, hence I have not provided narration on quantum and compound level consciousness.

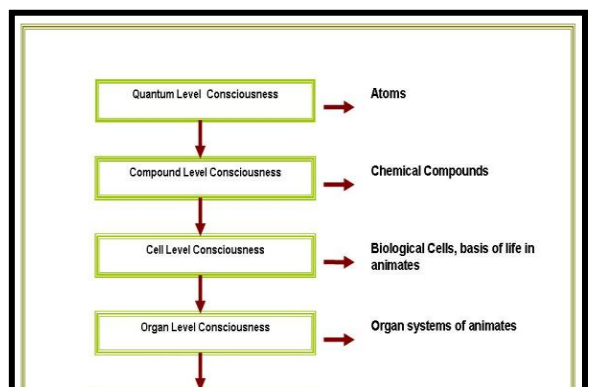

### 1.2.4 Illusion/Maya

This is the layer just below consciousness, that hides the complexity of framework of creation from thought/chinta layer. Hence, thought process of an animate matter do not requires to realize about consciousness or above that layer for it's normal functioning. **Illusion layer also hides simple principles of *Prakriti* and deludes as a complex phenomenon.**
**This layer is responsible for creation of various emotions such as: happiness, sad, anger, affections, infatuations etc.**
**In brief, this layer creates the illusions of duality in the nature which is not true in fact, such as pain and pleasure, affection and aversion, happiness and sorrow etc.,**

### 1.2,5 Thought / Chinta

This is the layer which creates procedures for actions that an animate performs while executing various task. **Prior to execution of every task, first a procedure of how to perform and what to perform is generated in the thought space**. Normally thought process is hidden from the complexity of creation through the duality created by illusion layer.

### 1.2.6 Stimulations/ Prayarthan

This is the layer that stimulates various organs of the body of animate according to the procedure generated at thought process to perform actions / karma. When stimulations crossed a certain threshold level, then only action is generated

### 1.2.7 Action/Karma

Action is generated through the stimulation/ prayarthan according to the procedure generated at thought space.

> **Action/Karma = Thought or Chinta + Stimulation**

## 2. Study of Consciousness
### 2.1 Definition of Consciousness

Consciousness can be logically defined as a function of parasitic, symbiotic, self referral and reproductive behavior.
Consciousness consists of following characteristics:
1. Parasitic Behavior,
2. Symbiotic Behavior,
3. Capacity to Refer Self, and
4. Reproductive Behavior.

**Parasitic Behavior**

It is the conscious behavior or tendency to acquire resources for self survival without concerning other's existence.
This behavior makes a being autonomous, and generates the tendency to compete for resource for self survival.

### 2.1.2 Symbiotic Behavior

It is the conscious behavior or tendency to associate peers for strengthening and smoothening the survival.
This behavior generates the social behaviors and allows the being to live in a social ambience.

### 2.1.3 Capacity To Refer Self

It is the process of recursively referencing self and generating actions with respect to self.
This behavior allows to distinctly positioning self and performing actions with respect to self and creates the capability to identify self with respect to the surrounding and acquiring resources for self survival autonomously.

### 2.1.4 Reproductive Behavior

Here, Reproduction means, reproducing a new combination from existing set of already produced objects. Here object means the characteristics traits.
This behavior or characteristics is used to create a new combination from existing set of characteristics. The resultant combination may be a biological child or a thought created inside brain that generates an intelligent action as a response to an external stimuli.
Above characteristics are essential yardsticks to have a heuristic measure of consciousness, which is again essential to test and implement consciousness.

Parasitic, Symbiotic, self referral and reproductive are the essential properties to categorize a specific entity as conscious.

## 3. Architecture of Consciousness

Architecture of consciousness consists of the following layers in existing conscious being.
1. Cell level,
2. Organ level,
3. Behavioral level.

### Cell Level Consciousness

The **cell** is the structural and functional unit of all living organisms, and are called as the "building blocks of life".

### 3.1.1 Test of consciousness at Cell level

1. Each cell is autonomous in taking input without regarding the existence of other cell(Parasitic behaviour)
2. Cells of same type joins together to form an organ(Symbiotic behavior)
3. Cell have reproductive capablity to create a another cell of same type(Reproductive Behavior).
4. Cell can exist independently and can collect resource for self existence(Self Referal).

Hence, it can be established that cell is the smallest unit of consciousness in a conscious architecture of a conscious being.

### Organ level consciousness

Similar cells are symbiotically grouped to form an organ and several organs symbiotically connected to form the complete physical skeleton as well as instructional skeleton. Each organ's functionality is distinct and uniquely instructed to perform an unique and specialized functionality. Each organ behaves parasitically to compete for resources needed for survival of the organ. Each organ grouped symbiotically to form the complete physical as well as instructional skeleton. Each organ have the capability to reproduce the cells to keep intact and grow the organ. Hence, Organ level consciousness exists.

### 3.2.1 The Comparison of cell level consciousness and organ level consciousness

Cell level consciousness are integration of differently instructed organelles in side cell. Where as organ level consciousness are integration of differently instructed cells in the organ. Cell have all the symptoms of consciousness. Organ level consciousness also have all the symptoms of consciousness derived through the cells present in that organ.

### 3.3 Behavioral level Consciousness

Here we regard behavioral consciousness is a complete autonomous system that consists of several organs and can communicate with the external world. Consciousness at cellular level and at organ level worked within the closed system and for the closed system, but here the complete body have consciousness to interact with the external system. Cells are integrated to from organs and organs are integrated to form the complete conscious system that can now called a conscious being.

### 3.3.1 How Consciousness at cell level and organ level are distinguished from that of consciousness at body level

- Consciousness at cell level or at organ level are, they worked at a functional level but consciousness at body level work at behavioral level, though heads of symptoms of consciousness are same.
- In this case nervous system plays a critical role to integrate rest all organs and produce conscious behavior externally.

A conscious being exhibits parasitic behavior to the external world by competing for resources for self survival, symbiotically form the social ambience for smoothening it's survival, Self referral by performing all actions with respect to self, and reproduce the self species type to populate and strengthening the species survival.

## 4. Design of Conscious Machine Architecture

A conscious machine architecture can be made possible, by the following three layers:
1. Building blocks of Lowest level of architectures should be conscious, such as: Bits, Arithmetic Units etc.,. Here

conscious means, processing capability should be inducted at lowest level as well.
2. Component level consciousness should also be achieved by inducting *Consciousness Algorithm.*

3. High level interface should be designed for interacting with external world by incorporating various *Behavioral Algorithms*.

### Conscious Algorithm

Conscious algorithm has four categories
A. Parasitic Algorithm: This algorithm will allow the machine to acquire resource for self and exhibit parasitic behavior, essential for autonomous survival.
B. Symbiotic Algorithm: This algorithm category will allow the machine to establish social interactions with peer group autonomously.
C. Self Referral Algorithm: This algorithm category will allow the machine to behave mime characteristics.
D. Reproductive Algorithm: This category of algorithm will allow the machine to organize a suitable set of instructions from existing set of instructions that can intelligently respond to an external stimuli.

### Behavioral Algorithm

Behavioral algorithm will be used for inducing intelligent behavior into the machine and made it possible for the machine to sense and respond to the external world intelligently.

Various behavioral algorithms are:
1. Design algorithm for control processor
2. Designing the algorithm for implementing the event processes such as:
    a) Experience
    b) Learning
    c) Inheritance of generic characteristics
3. Designing the algorithm for implementing sense via various sensory components.
4. Designing the algorithm for recursive sensory (Or, self diagnose).
5. Designing algorithm to create an active virtual space(or, thought space)
6. Designing the algorithm for analyze the sensory inputs with respect to
    a) Active virtual space(thought space)
    b) Active personality space (Personality trait).
    c) Intensity of the current sensory input.
7. Designing the protocol algorithm to store and organize sensory inputs in the layered memory.
8. Algorithm for developing creativity.
9. Algorithm for intention and desire generation.
10. Algorithm for developing machine to machine social interaction.
11. Algorithm for design for social language development through interaction.

## 5. Conclusion & Future work

This paper provides the basis for designing a conscious machine. First it presents the study and analysis of the concept of consciousness in an existing biological conscious being, architecture of consciousness and then reproducing the same architecture in the machine to develop consciousness in the machine. The prime objective to develop consciousness is to reduce the dependency of human for creating and developing intelligence in the machine. The so proposed work can be enhanced by developing consciousness algorithm and behavioral algorithms. The next work includes reducing the algorithm into machine implementation form in a suitable machine architecture using suitable computer language.